\documentclass[10pt,twocolumn,letterpaper]{article}

\usepackage[pagenumbers]{cvpr} 
\usepackage{xcolor}

\usepackage[htt]{hyphenat}
\usepackage[most]{tcolorbox}
\usepackage{helvet}
\usepackage{enumitem}

\newtcolorbox{surveybox}[1][]{
  colback=gray!5,      
  colframe=gray!50,    
  title={\textbf{User Study Questionnaire}},
  fonttitle=\sffamily,
  fontupper=\sffamily\small, 
  #1
}

\usepackage{graphicx}
\usepackage{booktabs, multirow} 
\definecolor{cvprblue}{rgb}{0.21,0.49,0.74}
\usepackage[pagebackref,breaklinks,colorlinks,allcolors=cvprblue]{hyperref}

\def\paperID{82} 
\def\confName{CVPR}
\def\confYear{2026}

\definecolor{meta-pink}{RGB}{231, 28, 115} 

\title{Glove2Hand: Synthesizing Natural Hand-Object Interaction \\ from Multi-Modal Sensing Gloves}

\author{
    Xinyu Zhang$^{\dagger, \ddagger}$ \quad Ziyi Kou \quad Chuan Qin \quad Mia Huang \quad Ergys Ristani \\
    Ankit Kumar \quad Lele Chen \quad Kun He \quad Abdeslam Boularias$^{\ddagger}$ \quad Li Guan
    \vspace{0.8em} \\
    \scalebox{0.9}{Meta Reality Labs \qquad $^{\ddagger}$Rutgers University} \qquad 
   \scalebox{0.85}{
   \href{https://mlzxy.github.io/glove2hand}{\color{meta-pink}{mlzxy.github.io/glove2hand}}}
}

\begin{document}

\twocolumn[{%
\renewcommand\twocolumn[1][]{#1}%
\maketitle
\begin{center}
\centering
\includegraphics[width=0.9\linewidth]{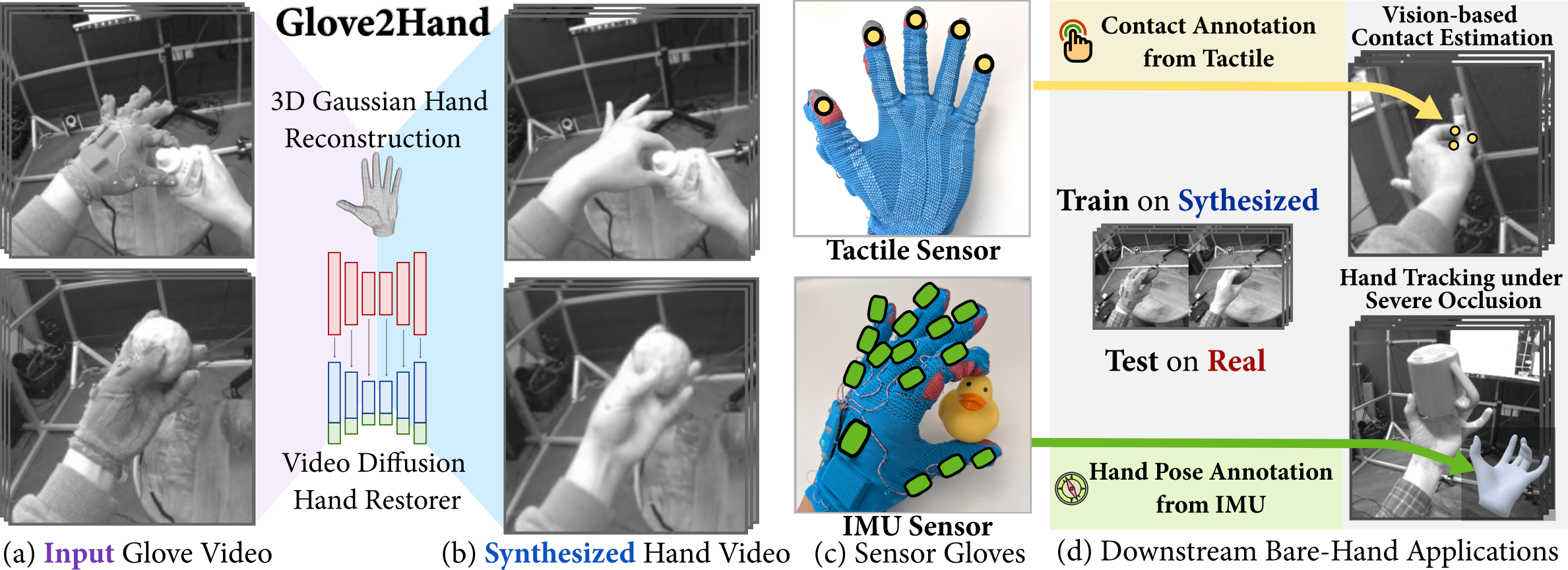}
\captionof{figure}{Glove2Hand translates egocentric glove videos (a) into photorealistic, temporally consistent, bare-hand videos (b) capturing complex interactions with non-rigid objects.
The translated videos are accompanied with synchronized tactile and IMU signals (c) from sensing gloves, which significantly enhances downstream bare hand tasks (d) by providing contact events from tactile as ground-truth for vision-based contact estimation; and occlusion-free hand poses from IMU as ground-truth for hand tracking under severe occlusion.
}
\label{fig:teaser}
\end{center}}]

\let\thefootnote\relax\footnotetext{$^{\dagger}$Work done during an internship at Meta Reality Labs.}

\begin{abstract}

Understanding hand-object interaction (HOI) is fundamental to computer vision, robotics, and AR/VR. However, conventional hand videos often lack essential physical information such as contact forces and motion signals, and are prone to frequent occlusions. To address the challenges, we present Glove2Hand, a framework that translates multi-modal sensing glove HOI videos into photorealistic bare hands, while faithfully preserving the underlying physical interaction dynamics. We introduce a novel 3D Gaussian hand model that ensures temporal rendering consistency. The rendered hand is seamlessly integrated into the scene using a diffusion-based hand restorer, which effectively handles complex hand-object interactions and non-rigid deformations. Leveraging Glove2Hand, we create HandSense, the first multi-modal HOI dataset featuring glove-to-hand videos with synchronized tactile and IMU signals. We demonstrate that HandSense significantly enhances downstream bare-hand applications, including video-based contact estimation and hand tracking under severe occlusion.
\end{abstract}

\section{Introduction}
Understanding human hand-object interaction (HOI) is a critical problem in computer vision, robotics, augmented reality (AR), virtual reality (VR), and human-computer interaction~\cite{haria2017hand, guo2025human, singh2025hand, luo2025being, huang2025hoigpt,carfi2021hand, mangalam2024enhancing,hao2024multimodal}. 
One common approach is to capture egocentric videos that contain various human hands, objects, and their interactions for developing data-driven algorithms~\cite{grauman2022ego4d, hoque2025egodex, ma2024diff, xu2023egopca}.
However, most HOI capturing systems nowadays are fundamentally limited to vision-only modality. As a result, they i) lack the rich physical grounding of force and contact information of hands; and ii) suffer from  hand occlusion due to limited camera views, which makes robust hand tracking difficult~\cite{hampali2020honnotate}.

Several vision-based solutions have been proposed to address the above limitations of HOI videos.
For example, ContactPose \cite{brahmbhatt2020contactpose} estimates binary fingertip contact from reconstructed mesh. But it only works with pre-scanned rigid objects and is not capable of providing continuous force measurements. Multi-camera studio \cite{moon2020interhand2} is designed to mitigate the hand occlusion issue, but is impractical for in-the-wild capture due to complex setup and calibration. 
Recently, sensing gloves have emerged as a new wearable device that captures various sensor modalities (e.g., IMU \cite{li2025fsglove, wang2025vihand}, tactile \cite{luo2024adaptive, zhu2024wearable}) and can be accompanied with egocentric camera devices for multi-modal data collection.
Fig. \ref{fig:teaser}.a shows an example of a sensing glove with full-hand IMU and fingertip tactile sensors.
Nonetheless, significant appearance gaps between such sensing gloves and human bare hands lead to low generalizability of vision models trained on glove data to bare hand tasks.

To address these limitations, we propose Glove2Hand, a generative 3D reconstruction-based video  approach that visually translates the sensing gloves in captured egocentric videos to photorealistic bare hands while faithfully preserving valuable IMU and tactile signals obtained from the sensor gloves. There are two major gaps in existing work that need to be addressed in order to achieve photo-realistic quality videos: (1) maintaining temporal and multi-view consistency across frames instead of focusing on only static images~\cite{chen2025foundhand,lu2024handrefiner}; and (2) handling interactions with complex objects (including non-rigid objects with unknown shapes) instead of those with known and rigid shapes~\cite{zhang2024hoidiffusion,xu2023interdiff}.


Glove2Hand combines the strengths of 3D reconstruction for consistency and generative modeling for flexibility and photo-realism. We first build a minimalist 3D Gaussian hand model, which defines statistical surface distributions over a learnable canonical hand mesh. The design provides a strong geometrical prior, naturally enables relighting, and renders spatiotemporally consistent object-free hand videos. Subsequently, unlike previous methods that explicitly model object geometry~\cite{zhang2024hoidiffusion, banerjee2025hot3d}, we represent objects and background in the pixel domain, allowing for greater flexibility in handling objects with unknown or deformable shapes. We then train a diffusion-based hand restorer to seamlessly integrate the rendered hands into the scene, refining hand-object details and overall coherence. 

To demonstrate the effectiveness of Glove2Hand on enhancing bare-hand learning tasks, we further create a multimodal HOI video dataset \emph{HandSense} that contains accurate glove poses, realistically synthesized bare-hand videos, and time-synchronized sensors including IMU and tactile signals. In particular, we design two bare-hand tasks on HandSense (Fig.~\ref{fig:teaser}d): \textit{Vision-based Fingertip Contact Estimation} and \textit{Hand Pose Tracking under Heavy Occlusion}. We conduct extensive experiments to show that Glove2Hand effectively enhances the model performance on both tasks for bare hands. Our contributions are three folds: 
\begin{itemize}
    \item We propose the Glove2Hand framework for glove-to-hand video translation, which leverages a novel 3D Gaussian hand for consistent rendering, and a diffusion hand restorer for generative refinement of hand-object details. To our best knowledge, this is the first work to generate photorealistic and authentic hand-object interactions from sensor glove captures,  bridging the large appearance gap between the sensor glove and human hand.
    
    
    \item We evaluate the generative quality of our Glove2Hand on our new dataset \textit{HandSense}, the first HOI dataset with synchronized bare-hand and sensor glove videos, and directly measured continuous tactile and IMU signals.
    
    \item We demonstrate the value of synthesized hand videos on two novel applications: Vision-based contact estimation and robust hand tracking under severe occlusion.
\end{itemize}

\begin{figure*}
  \centering
   \includegraphics[width=0.9\linewidth]{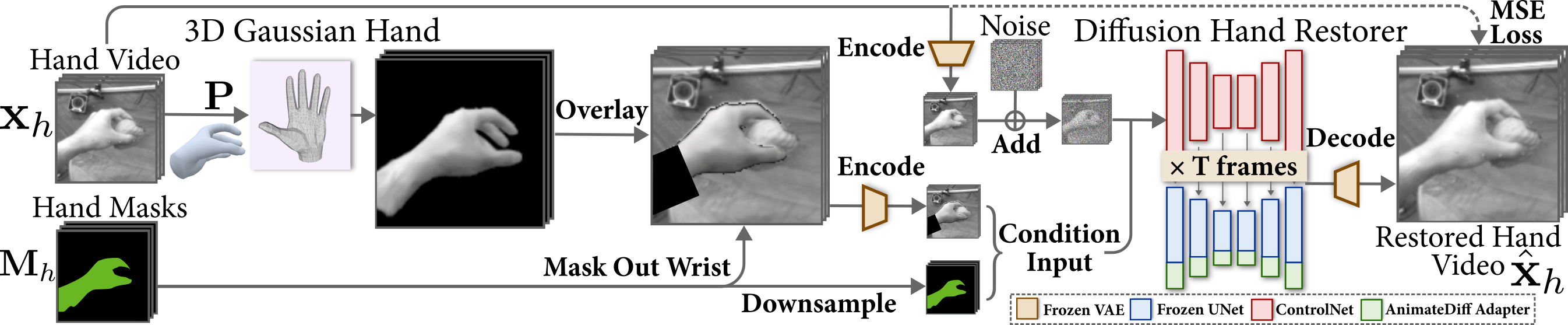}
   \caption{\textbf{Glove2Hand Training.} 
   We extract hand poses from input videos and render corresponding hand only frames using our 3D Gaussian hand model. The rendered hands are then overlayed onto the original videos with wrist regions masked. The resulting video is encoded into a latent space and serves as a condition for our diffusion hand restorer, which is trained to restore the original hand video.}
   \label{fig:method-train}
\end{figure*}

\begin{figure}
  \centering
   \includegraphics[width=0.9\linewidth]{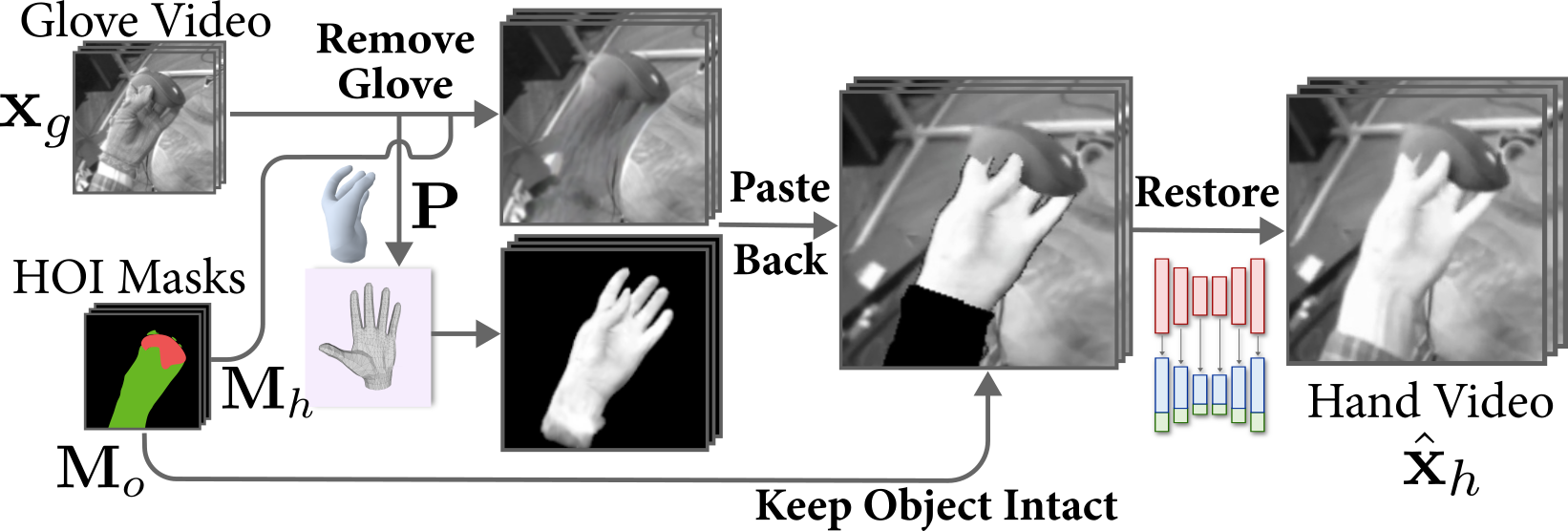}
   \caption{\textbf{Glove2Hand Inference.} We render hand only frames using hand poses from glove videos. Prior to pasting, we erase the glove region  with an optical-flow based background inpainter~\cite{zhou2023propainter}, and ensures object pixels remain intact during pasting. The resulting video is encoded and fed as input to diffusion hand restorer.}
   \label{fig:method-inf}
\end{figure}

\section{Related Work}

\textbf{Generative Models for Hands.} Research on generating hand imagery falls into three categories. \textbf{Hand avatars}~\cite{iwase2023relightablehands, dong2025handsplat, zheng2024ohta, karunratanakul2023harp, zhao2024gaussianhand, chen2023hand} use rendering methods like raytracing~\cite{braunig2023realistic}, NeRF~\cite{mildenhall2021nerf} or Gaussian Splatting~\cite{kerbl20233d, braunig2023realistic} 
to offer precise control, but typically generate only the hand without backgrounds or interaction.
\textbf{Hand diffusion models}~\cite{chen2025foundhand, pelykh2024giving, lu2024handrefiner, narasimhaswamy2024handiffuser, park2024attentionhand, zhang2025hand1000, MANO} generate plausible hands within general image synthesis pipelines, with backgrounds, but lack precise control and focus on single, low-detail, static images. Finally, \textbf{hand-object interaction synthesis}~\cite{ye2023affordance, xu2023interdiff, zhang2024hoidiffusion} specializes in generating close-up, static images of contact, for instance by estimating grasp poses from object images~\cite{ye2023affordance} or rendering from pre-scanned meshes~\cite{zhang2024hoidiffusion}.

Our framework distinguishes from these approaches in two key aspects. First, we generate dynamic \textbf{videos} of hand-object interaction, capturing actions rather than the static images produced by prior work. Second, the generated hand motions are \textbf{grounded by an input glove video}. This grounding provides precise controllability, enabling multi-modal sensor fusion and scalable data curation without requiring hard-to-obtain non-rigid object meshes.

\noindent\textbf{Video-to-Video Translation.} Early image-to-image translation learned mappings from aligned~\cite{isola2017image} or unpaired data~\cite{zhu2017unpaired}, but cannot handle the large  embodiment differences in our task. Recent video translation work~\cite{yang2023rerender, hu2023videocontrolnet, yang2024fresco, chu2024medm, bian2025videopainter} often relies on latent space interpolation in pre-trained image diffusion models. In contrast, our method learns a direct glove-to-hand video translation from unpaired data, addressing a significant domain gap without being constrained by a pre-trained model's capabilities.

\noindent\textbf{Video Inpainting.} Video inpainting traditionally focused on object removal and background completion, often using optical flow for temporal consistency~\cite{zhou2023propainter, li2025diffueraser}. More recent methods leverage diffusion models for language-conditioned inpainting~\cite{ju2024brushnet, yu2023inpaint, gu2024coherent}. Our work is distinct in its focus on inpainting highly detailed hand-object contact regions, a particularly challenging sub-problem due to the dexterity of hands and the complexity of contact physics.

\noindent\textbf{Hand Datasets with Tactile Sensing.} Datasets for hand-object interaction~\cite{banerjee2025hot3d, liu2022hoi4d, wang2024ho} often estimate binary contact from meshes~\cite{brahmbhatt2020contactpose, kwon2021h2o, fan2023arctic}. While some provide continuous force, they are often limited to flat tablets (restricting dexterity)~\cite{grady2022pressurevision, grady2024pressurevision++, zhao2025egopressure, senselpad}, are simulation-based~\cite{xu2023visual}, or are glove-only datasets lacking visual data~\cite{jiang2025kaiwu}. To our knowledge, our work creates the first dataset combining sensor-measured tactile and IMU signals with synchronized, photorealistic bare-hand videos of dexterous manipulation.

\section{Glove-to-Hand Framework}

\subsection{Problem Formulation}

We define a video clip as $\mathbf{x} \in \mathbb{R}^{T \times H \times W \times C}$ in either glove domain $\mathbf{x}_g \in \mathcal{I}_{\text{glove}}$ or hand domain $\mathbf{x}_h \in \mathcal{I}_{\text{hand}}$. Each clip is associated with hand poses $\mathbf{P} \in \mathbb{R}^{T \times J \times 3}$, hand and object masks, $\mathbf{M}_h, \mathbf{M}_o \in \{0,1\}^{T \times H \times W}$, forming a sample tuple $s = (\mathbf{x}, \mathbf{P}, \mathbf{M}_h, \mathbf{M}_o)$. Let $\mathcal{D}_{\text{glove}} = \{s \mid \mathbf{x} \in \mathcal{I}_{\text{glove}}\}$, our objective is to learn a mapping $f: \mathcal{\mathcal{D}_{\text{glove}}} \mapsto \mathcal{I}_{\text{hand}}$ that translates a glove sequence into a photorealistic hand video without manual edits. We assume $\mathcal{I}_{\text{glove}}$ and $\mathcal{I}_{\text{hand}}$ are unpaired, where no sample correspondence exists across domains.

\subsection{Method}

The key challenges in realistic glove-to-hand video translation are twofold: (1) achieving temporal and multi-view consistency, which is notoriously difficult for generative models, and (2) modeling complex interactions with non-rigid objects. This is further compounded by the significant visual domain gap between the sensor glove and a bare hand. Our core insight is that despite this appearance gap, both the glove and hand share the same underlying articulated structure, represented by the hand pose $\mathbf{P}$. 

This realization allows us to decompose the problem into two more tractable subproblems: (1) transforming the glove video into a consistent, in-air hand sequence, and (2) integrating this hand sequence back into the scene while refining interaction details. Therefore, we combine the consistency of a reconstruction-based hand avatar (to solve the first subproblem) with the generative flexibility of a diffusion model (to solve the second). This approach allows us to refine complex hand-object boundary details without requiring explicit 3D representations of objects or the background. Our framework is visualized in Fig.~\ref{fig:method-train} and Fig.~\ref{fig:method-inf}.

\subsubsection{3D Gaussian Hand}
\label{sec:3d-gs-hand}


Hand avatar methods deterministically reconstruct 3D hand representations and render novel views from a given hand pose and camera pose, typically using neural rendering techniques rather than generative sampling. However, applying existing avatar methods to our scenario presents several challenges: (1) state-of-the-art methods often require dense, multi-view camera setups (e.g., hundreds of views~\cite{moon2020interhand2,iwase2023relightablehands}), whereas our egocentric headset provides only two closely positioned cameras; (2) existing avatar datasets are typically captured under controlled, calibrated lighting conditions. In contrast, our egocentric scenario features frequent and dynamic lighting changes inherent to real-world human movement. Ignoring lighting variations leads to ambiguous reconstructions; (3) established methods often rely on heavy, multi-stage pipelines involving NeRF or ray marching~\cite{mundra2023livehand, iwase2023relightablehands, chen2023hand, karunratanakul2023harp}.
While recent work has explored Gaussian Splatting for hand avatars~\cite{dong2025handsplat, zhao2024gaussianhand, sun2025jghand}, their implementations are complex and not publicly available. As this avatar serves as a component within our larger framework, we prioritize simplicity and efficiency. We therefore develop a minimalist 3D Gaussian hand model adapted for sparse camera inputs and variable lighting.

\begin{figure}
  \centering
   \includegraphics[width=0.9\linewidth]{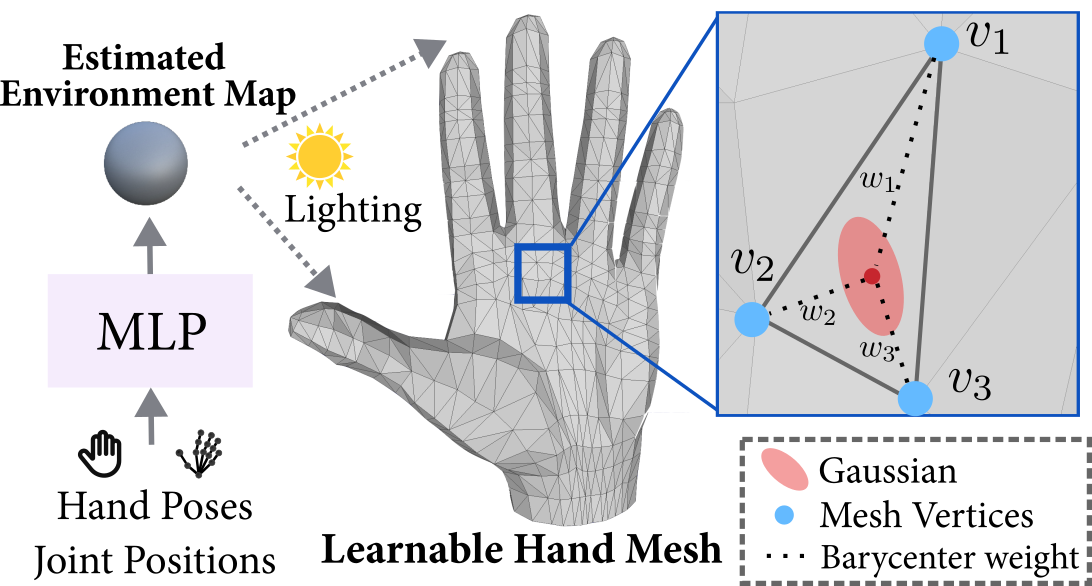} 
   \caption{\textbf{Surface-grounded and Relightable Gaussian Hand.} We defines Gaussians over the surface of a hand mesh, which benefits from the mesh geometric priors and naturally enables lighting estimation based on the consistent mesh surface normals. 
   } 
   \label{fig:gaussian-hand}
\end{figure}

Our design is motivated by a key insight: Canonical hand meshes provide a strong geometric prior but lack learning flexibility, whereas Gaussian splatting is highly flexible but lacks inherent structure. We unify these advantages by defining 3D Gaussians directly on the canonical mesh surface. This simple yet effective modification also naturally addresses the challenge of variable lighting. The underlying mesh provides consistent surface normals which are essential for enabling robust lighting estimation. Our two proposed techniques are detailed below and illustrated in Fig~\ref{fig:gaussian-hand}.

\noindent\textbf{Surface-Grounded Gaussians.} We define Gaussians directly on mesh surfaces. For a triangle with vertices $v_1, v_2, v_3$, each Gaussian is characterized by barycentric weights $\mathbf{w} = [w_1, w_2, w_3]$ where $\sum_{i} w_i = 1$, 2D scale $\mathbf{s} = [s_x, s_y]$, and rotation $\phi$. Unlike 2DGS~\cite{huang20242d} which defines Gaussians in 3D space with regularization to encourage surface formation, our method leverages a stronger prior and directly anchors Gaussians to mesh surfaces.

During hand deformation, instead of applying linear blend skinning to Gaussians (which can break surface integrity~\cite{zhang2025motion}), we transform only the mesh triangles and recompute Gaussians on the deformed surfaces. This leverages the hand mesh's pre-existing bone weights, avoiding the learning of skinning weights for individual Gaussians.

Formally, given triangle vertices $v_1, v_2, v_3$ in canonical frame and $v'_1, v'_2, v'_3$ in deformed frame, we compute edges $\mathbf{e}_1 = v_2 - v_1$, $\mathbf{e}_2 = v_3 - v_1$ and similarly $\mathbf{e}'_1 = v'_2 - v'_1$, $\mathbf{e}'_2 = v'_3 - v'_1$ for the deformed frame. These edges are projected to local 2D coordinates using an orthonormal basis $(\mathbf{u}, \mathbf{v})$ derived via Gram-Schmidt: $\mathbf{u} = \frac{\mathbf{e}_1}{\|\mathbf{e}_1\|}$, $\mathbf{w}_\perp = \mathbf{e}_2 - \text{proj}_\mathbf{u}(\mathbf{e}_2)$, $\mathbf{v} = \frac{\mathbf{w}_\perp}{\|\mathbf{w}_\perp\|}$, yielding $\mathbf{e}_{1,D} = (\mathbf{e}_1 \cdot \mathbf{u}, \mathbf{e}_1 \cdot \mathbf{v})$ and $\mathbf{e}_{2,D} = (\mathbf{e}_2 \cdot \mathbf{u}, \mathbf{e}_2 \cdot \mathbf{v})$.
We point out that any point $p$ on the triangle can be expressed as $p = c_1\mathbf{e}_{1,D} + c_2\mathbf{e}_{2,D}$, and the coefficients $\mathbf{c} = [c_1, c_2]^T$ shall remain constant under deformation. This gives us $p = \mathbf{M}_{\text{canon}}\mathbf{c}$ and $p' = \mathbf{M}_{\text{deform}}\mathbf{c}$ for the deformed frame, where $\mathbf{M}_{\text{canon}} = [\mathbf{e}_{1,D}; \mathbf{e}_{2,D}]$ and $\mathbf{M}_{\text{deform}} = [\mathbf{e}'_{1,D}; \mathbf{e}'_{2,D}]$. The deformation gradient $\mathbf{A} = \mathbf{M}_{\text{deform}}\mathbf{M}_{\text{canon}}^{-1}$ maps point from canonical frame to deformed frame by $p' = \mathbf{A}p$.
A surface-grounded Gaussian can be represented by an ellipse in quadratic form $\mathbf{Q} = \mathbf{R}\mathbf{S}\mathbf{R}^T$. $\mathbf{R}$ is the 2D rotation matrix for $\phi$ and $\mathbf{S} = \text{diag}(1/s_x^2, 1/s_y^2)$. The deformed ellipse becomes $\mathbf{Q}' = \mathbf{A}^{-T}\mathbf{Q}\mathbf{A}^{-1}$, from which new scale $s'$ and rotation $\phi'$ are obtained via eigen decomposition.

Each Gaussian is associated with a triangle. The adaptive control algorithm can be directly applied to flexibly allocate Gaussians—during each control cycle, we reassign Gaussians to their closest triangles. We learn a small per-Gaussian offset along the surface normal, ensuring overlapping Gaussians do not share the same depth, thereby avoiding ambiguity in point-based rendering~\cite{gross2011point}.
We train these Gaussians using only image reconstruction loss through differentiable rendering. This hierarchical definition allows gradients to jointly finetune the canonical mesh vertices.

\noindent\textbf{Relightable Hand Gaussians.}
To model variable illumination, we adapt the relighting framework proposed by LumiGauss~\cite{kaleta2025lumigauss}. This approach learns an environment map, represented as spherical harmonics (SH) coefficients $\mathbf{l} \in \mathbb{R}^{3 \times (n^2+1)}$, where $n$ is the SH order. The final color is computed as $\mathbf{c} \odot \text{SH}(\mathbf{l}, \mathbf{n})$, where $\mathbf{c}$ is the Gaussian's intrinsic color (albedo) and $\text{SH}(\mathbf{l}, \mathbf{n})$ evaluates the diffuse lighting component given the surface normal $\mathbf{n}$. However, this method presents two limitations for our dynamic, egocentric setting: (1) LumiGauss assumes a single static environment map, whereas our lighting conditions change dynamically;  (2)  it suffers from the well-known albedo-illumination ambiguity. As noted in LumiGauss~\cite{kaleta2025lumigauss}, the learned Gaussian normals $\mathbf{n}$ and intrinsic colors $\mathbf{c}$ can co-adapt to incorrectly bake shadows into the albedo.

We address both limitations. To solve the first, we model lighting as a dynamic function of the hand's configuration. We employ a small MLP to predict the SH coefficients $\mathbf{l}$ from hand pose $\mathbf{P}$. This is effective because local lighting and self-shadowing are primarily determined by the hand's location and articulation. To solve the second, our core design of defining Gaussians on the mesh provides a natural solution. The surface normals $\mathbf{n}$ are provided by the mesh geometry but not Gaussian, which substantially mitigates the ambiguity. In practice, to better capture local effects, our MLP predicts two separate environment maps, $\mathbf{l}_{\text{palm}}$ and $\mathbf{l}_{\text{back}}$, for the palm and back of the hand, respectively.

\subsubsection{Diffusion Hand Restorer}
\label{sec:diff-hand-restore}

Our 3D Gaussian hand renders a consistent, hand-only sequence from the glove video. However, naively overlaying the render introduces several artifacts, such as implausible object interactions like penetration or floating, as the render is not interaction-aware; and an unnatural wrist connection, as the arm is not modeled. Some of these artifacts are illustrated in Fig.~\ref{fig:ablation}. Furthermore, the bulky glove's larger shape, when replaced by the slimmer hand, leaves visible gaps around the hand.
To address these limitations, we introduce a diffusion hand restorer based on ControlNet~\cite{zhang2023adding} and AnimateDiff~\cite{rombach2021highresolution}, and trained on bare-hand videos. We generate the conditional input by overlaying the rendered hand, dilating its mask, and masking the wrist region. The network is trained to restore the original bare-hand video from such corrupted input, as shown in Fig.~\ref{fig:method-train}.


\noindent\textbf{Glove-to-Hand Inference.} Gloves  have a different and larger shape than hands. Therefore, overlaying a rendered hand onto  glove videos creates visible glove artifacts around the rendered hand's edges (illustrated in Fig.~\ref{fig:ablation}, columns 2-4), and can also incorrectly occlude foreground objects.
To resolve this, we employ a glove-to-hand inference pipeline (Fig.~\ref{fig:method-inf}).
Specifically, we first extract glove and object masks using SAM-2~\cite{ravi2024sam}, prompted by bounding boxes detected with Grounding DINO~\cite{liu2024grounding}. Next, we use the glove mask to erase the glove region with Propainter~\cite{zhou2023propainter}, an optical flow-based background inpainter. We leverage the object mask to ensure that foreground object pixels remain intact. The entire pipeline is fully automatic.

\section{HandSense Dataset}
To validate the effectiveness of Glove2Hand, we create \textit{HandSense}, a multi-modal HOI video dataset that contains egocentric videos with time-synchronized sensor signals. 
In particular, we recruited 5 subjects and performed two sessions of data collection from each. In this first session, the subject was asked to wear the sensing glove and complete 6 object manipulation tasks (e.g., in-hand rotation, pick-and-place).
In the second session, the subject followed the same object manipulation protocol but didn't wear the gloves.
We show the collected data in Figure \ref{fig:dataset} and more data collection details in the supplementary materials.

\noindent\textbf{Data Modalities.} For both sensing glove and bare hand sessions, we collect egocentric grayscale video recordings from headset cameras. To obtain accurate 3D hand poses, we adopt MoCap system \cite{ray2024comprehensive, vallaro2024markerless} with markers on the gloves/hands. For the sensing glove session, we collected IMU signals distributed on multiple hand phalanges, and tactile signals located at each fingertip. Both IMU and tactile signals are continuous values and time-synchronized  with egocentric videos based on the closest timestamps. For the bare hand session, we manually annotate discrete binary contact/no-contact labels for each fingertip when a subject is manipulating an object. To mitigate the hand occlusion issue, we carefully designed the instructions for subjects to use specific fingers so that we have strong prior knowledge about contact events.


\begin{figure}
  \centering
   \includegraphics[width=0.9\linewidth]{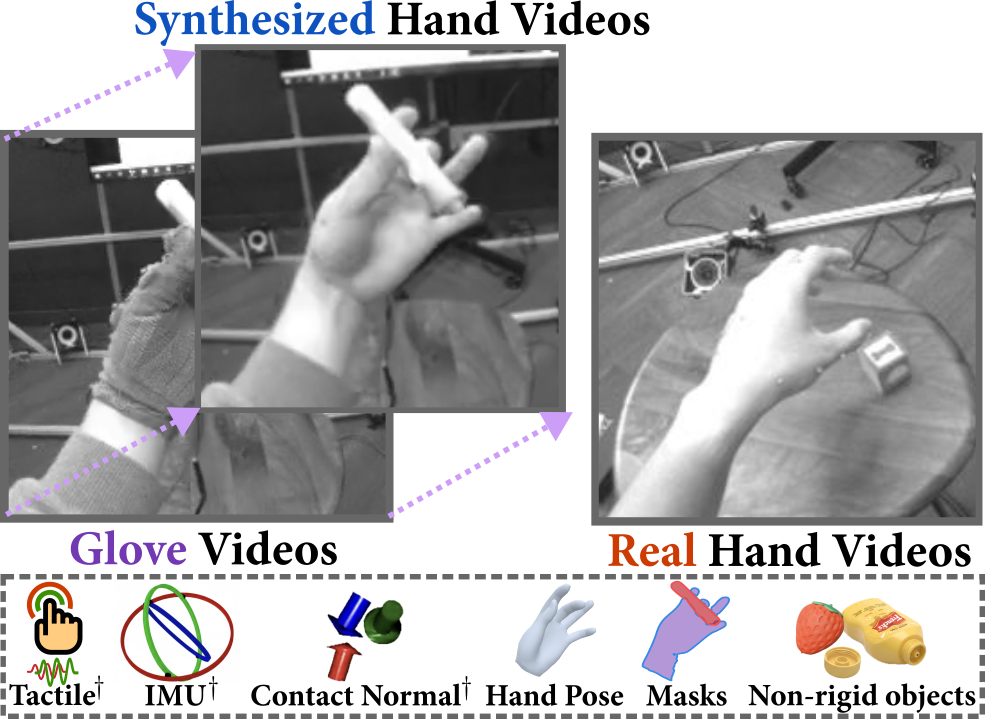}
   \caption{\textbf{HandSense Dataset.} HandSense provides egocentric HOI videos collected by both sensing gloves and human bare hands. $\dagger$ denotes  multi-modal sensing modalities from gloves.
   }
   \label{fig:dataset}
\end{figure}


\noindent\textbf{Measured Tactile Signals and Non-rigid Objects.} We hightlight two key features that distinguish HandSense from existing datasets. First, to our best knowledge, HandSense provides the first continuous, sensor-measured tactile signals synchronized with bare-hand HOI videos. In contrast, datasets like H2O~\cite{kwon2021h2o} and ARCTIC~\cite{fan2023arctic} provide only binary, post-processed contact labels, which are estimated from hand and rigid object poses rather than measured. Second, HandSense features interactions with diverse object categories, including deformable (e.g., squishy toys) and articulated items (e.g., detachable lids), unlike prior methods that are often limited to pre-scanned, rigid objects.

With HandSense dataset, we demonstrate the value of Glove2Hand in two bare-hand applications: i) video-based contact estimation, benefited from 
tactile signals by sensing gloves as annotation-free ground truth, and ii) robust hand tracking under occlusion that augments estimated hand poses in occluded scenarios using IMU signals as pseudo ground-truth. 
The objective of both tasks is to facilitate deeper physical understanding of hand-object interaction based on sensor signals from gloves. We show more detailed experiment setups and results in Section \ref{sec:exp-app}.



\begin{figure*}[t]
  \centering
   \includegraphics[width=0.9\linewidth]{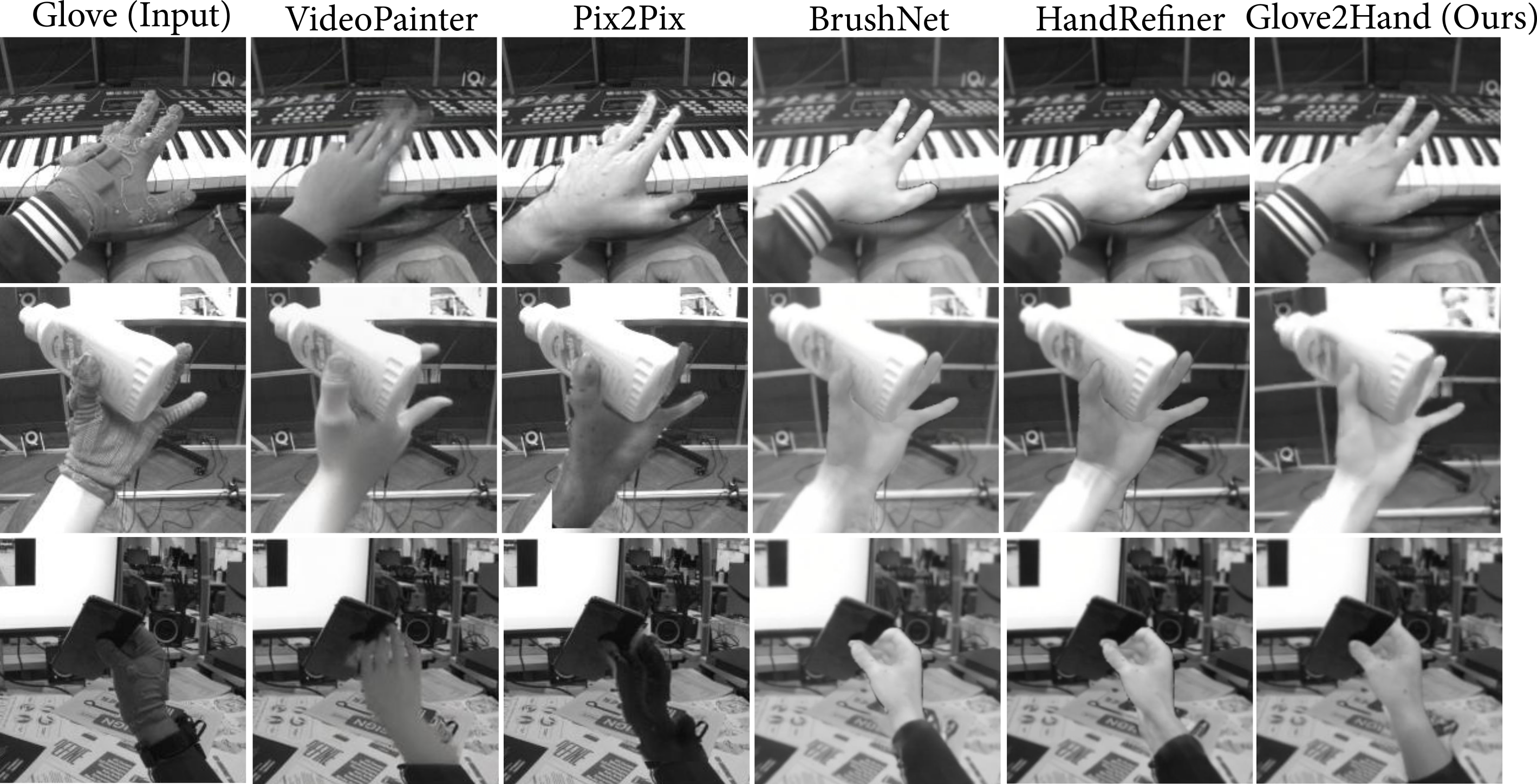}
   \caption{\textbf{Qualitative Comparison for Glove-to-Hand.} Our Glove2Hand produces photorealistic hands with clean object contact. In contrast, 
   existing methods generate blurry results, or suffer from strong visual artifacts, unrealistic textures, and inconsistent lighting. The bottom row demonstrates the high generalization of Glove2Hand to unseen subjects, objects, and backgrounds. For a fair comparison, Pix2Pix, BrushNet, and HandRefiner were retrained on our dataset using the same inputs and HOI masks as our method (Fig.~\ref{fig:method-train} and Fig.~\ref{fig:method-inf}).}
   \label{fig:qualitative}
\end{figure*}

\section{Experiments}
\label{sec:exp}

\begin{table}\centering
\begin{tabular}{lccc}\toprule
\textbf{Method} &\textbf{FID $\downarrow$ } &\textbf{FVD $\downarrow$ } &\textbf{FVD-long $\downarrow$ } \\\midrule
Vanilla Glove &74.9 &38.3 &45.5 \\
InpaintAnything$^*$~\cite{yu2023inpaint} &63.6 &32.7 &39.6 \\
FoundHand$^*$~\cite{chen2025foundhand} &82.2 &47.9 &50.5 \\
VideoPainter$^*$~\cite{bian2025videopainter} &72.2 &35.7 &42.2 \\
CycleGAN~\cite{zhu2017unpaired} &63.6 &30.8 &37.4 \\
Pix2Pix~\cite{isola2017image} &38.6 &24.7 &31.4 \\
BrushNet~\cite{ju2024brushnet} &37.9 &34.5 &40.4 \\
HandRefiner~\cite{lu2024handrefiner} &35.5 &24.2 &29.7 \\
Glove2Hand (Ours) &\textbf{30.1} &\textbf{19.5} &\textbf{24.5} \\
\bottomrule
\end{tabular}
\caption{\textbf{Evaluation for Glove-to-Hand Video Translation.} We evaluate Glove2Hand on HandSense dataset with FID, FVD (2-sec. clips), and FVD-long (on 10-30 sec. clips). 
($*$) denotes methods evaluated off-the-shelf without fine-tuning on HandSense. 
}
\label{tab:main}
\end{table}

\begin{figure}[t]
  \centering
   \includegraphics[width=0.9\linewidth]{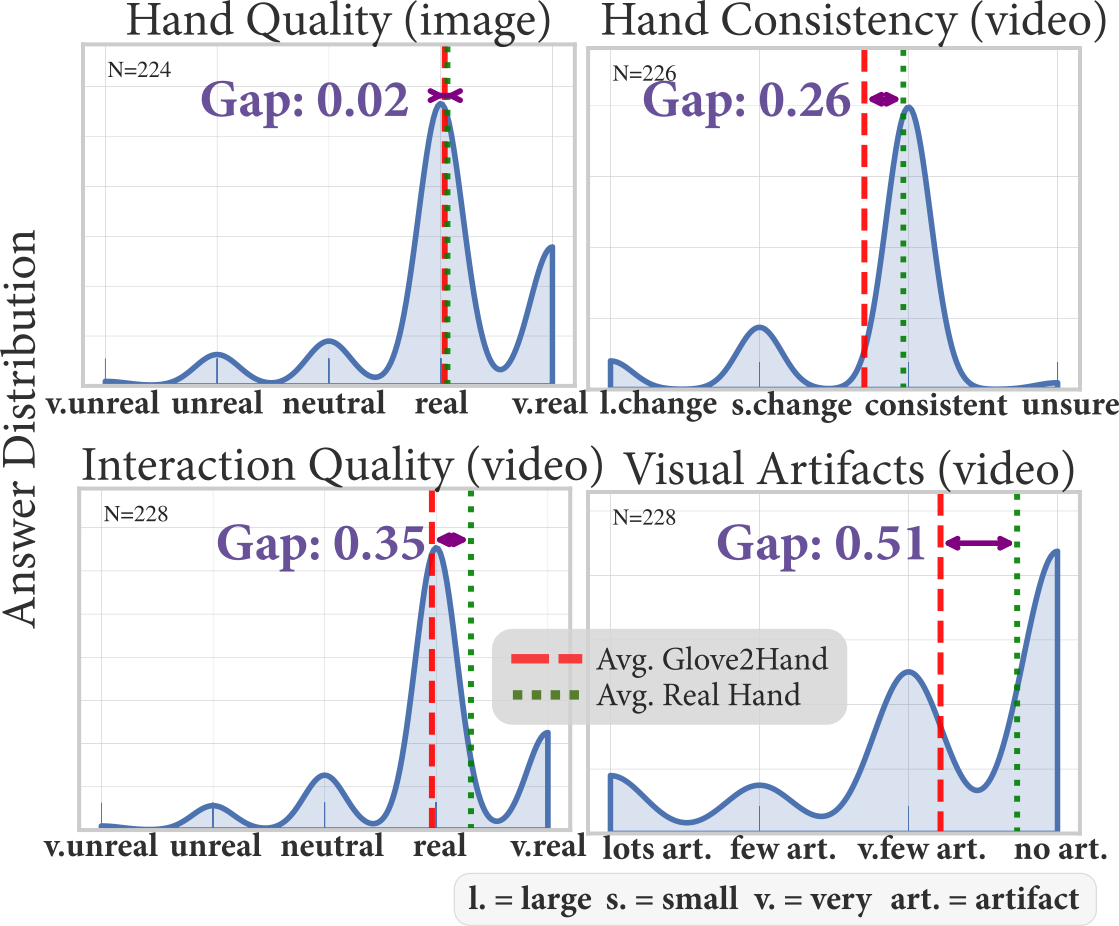}
   \caption{\textbf{Human Evaluation Results} . Our study  compares the synthesized videos by Glove2Hand with real hand videos across four metrics ($N$=number of answers). The average of score distribution for our synthesized hands are close to real hands, with a max gap of 0.51. This gap (lower is better) is small relative to the 1.0 distance between adjacent categories, showing high realism.}
   \label{fig:user-study-video}
\end{figure}

We design our experiments to answer three key questions:

\begin{enumerate}
    \item How realistic are the hand-object interaction (HOI) videos generated by our Glove2Hand framework?
    \item What is the contribution of each component of our proposed framework to the final generation quality?
    \item What is the value of our generated HOI data with synchronized sensor signals, for bare hand applications?
\end{enumerate}

To address the first, we evaluate our framework on HandSense using quantitative metrics and human evaluation. Second, we conduct ablations on each component. For the third, we demonstrate our data's value in  video-based contact estimation and hand tracking under heavy occlusion.

\noindent\textbf{Sensing Glove.} Our sensing glove includes one tactile sensor per fingertip which produces one channel of continuous force values, and twelve 6-DoF IMU sensors distributed on the full hand for tracking hand pose. We provide more details of the glove configurations in the Supplementary. 


\subsection{Video Quality Evaluation}
\label{sec:exp-vqe}

\begin{figure*}[t]
  \centering
   \includegraphics[width=0.9\linewidth]{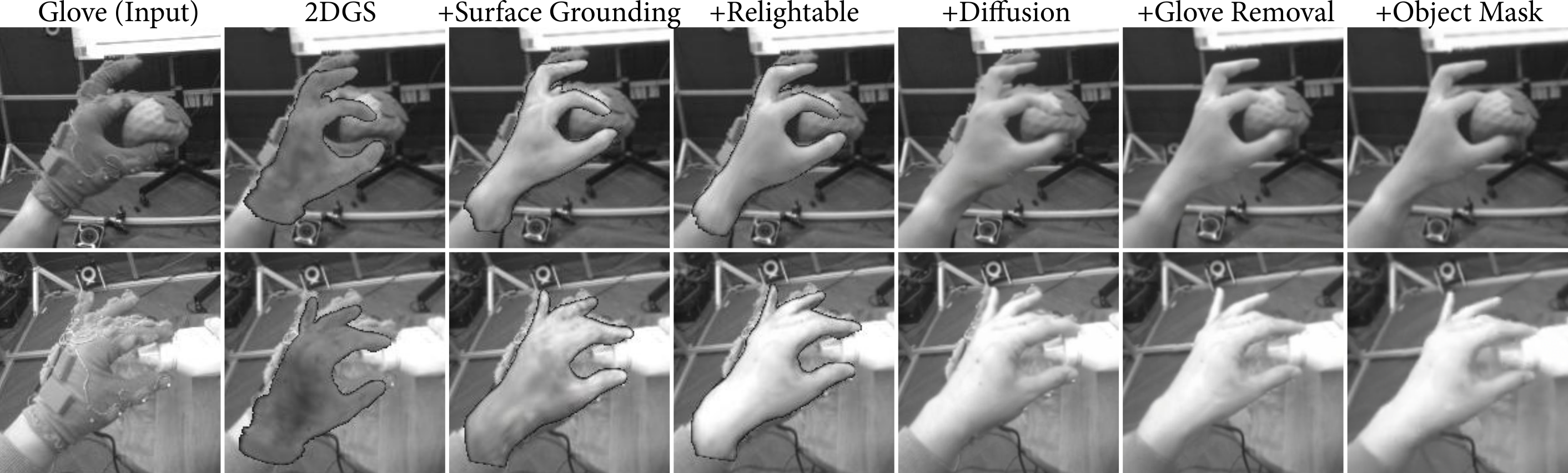}
   \caption{\textbf{Qualitative Ablations.} Our surface-grounded Gaussian reconstructs coherent hand geometry and appearance compared to 2DGS~\cite{huang20242d}. Adding the relightable Gaussian improves shading. The diffusion hand restorer further enhances skin texture, connects the wrist, and removes visual artifacts. Finally, glove removal and object masks eliminate glove artifacts and refine the hand-object boundary.}
   \label{fig:ablation}
\end{figure*}

\noindent\textbf{Setup.} We train our framework on the HOT3D~\cite{banerjee2025hot3d} and  HandSense datasets, and evaluate on the testing split of HandSense. We firstly optimize a 3D Gaussian hand model for each subject. The subject-specific models are then frozen while training the single diffusion restorer. We perform same-subject glove-to-hand translation by using the corresponding 3D Gaussian hand. We measure realism against real hand videos using Fréchet Inception Distance (FID)~\cite{heusel2017fid} (clean-fid~\cite{parmar2022aliasedfid}) for image quality and Fréchet Video Distance (FVD)~\cite{unterthiner2018fvd} (cd-fvd~\cite{ge2024cdfvd}) for video fidelity.

\noindent\textbf{Results.} We present quantitative and qualitative comparisons in Tab.~\ref{tab:main} and Fig.~\ref{fig:qualitative}. Glove2Hand significantly outperforms existing generative and vidoe inpainting baselines in both FID and FVD metrics. We omit some baselines from Fig.~\ref{fig:qualitative} as they fail on glove-to-hand translation by returning the input gloves largely unchanged. Compared to diffusion-only methods like HandRefiner~\cite{lu2024handrefiner}, our 3D Gaussian representation synthesizes finer geometry and texture.

\noindent\textbf{Human Evaluation.} We conduct a user study to assess the realism of our generated hand-object interaction videos. We recruited five participants and presented them with a mix of our generated and real hand videos. For each sample, subjects answered questions regarding hand realism, the plausibility of the hand-object interaction, temporal consistency (e.g., identity preservation), the presence of visual artifacts, and hand motion stability. Each participant evaluated approximately over 40 images and 40 videos. We visualize the aggregated response distributions in Fig.~\ref{fig:user-study-video}. The results demonstrate that our generated videos achieve a high degree of realism, approaching the scores of real videos. Notably, for still images, participant responses indicate that our generated hands are nearly indistinguishable from real ones. We provide further details in the Supplementary.

\begin{table}\centering
\begin{tabular}{lccc}\toprule
\textbf{Configuration} &\textbf{FID $\downarrow$} &\textbf{FVD $\downarrow$ } &\textbf{FVD-long $\downarrow$} \\\midrule
2DGS~\cite{huang20242d} &91.1 &50 &62.9 \\
+Surface Grounding &60.3 &35.1 &46.6 \\
+Relightable &56.7 &30.7 &40.2 \\
+Diffusion &32.3 &19.8 &\textbf{22.7} \\
+Glove Removal &31.2 &20.9 &25 \\
+Object Mask &\textbf{30.1} &\textbf{19.5} &24.5 \\
\bottomrule
\end{tabular}
\caption{\textbf{Ablation Results.} We perform an incremental ablation study to evaluate the contribution of each proposed component. The results demonstrate that our components work cohesively, with each addition progressively improving the final performance. Surface grounding and the diffusion yield the most significant gains. Qualitative ablations are presented in Fig.~\ref{fig:ablation}.}\label{tab:ablation}
\end{table}

\noindent\textbf{Ablation Studies.} We present quantitative and qualitative ablation studies in Tab.~\ref{tab:ablation} and Fig.~\ref{fig:ablation}, respectively. The quantitative results show that each component incrementally improves image fidelity (FID). While we observe minor FVD fluctuations, the qualitative results in Fig.~\ref{fig:ablation} clearly demonstrate the distinct and necessary contribution of each module. Specifically, our surface-grounded and relightable 3D Gaussian hand establishes a realistic, geometrically-accurate foundation. The Diffusion Hand Restorer builds upon this, fine-tuning skin textures, generating a natural wrist, and refining hand-object boundaries. The glove removal module is critical for eliminating visual artifacts from the input sensor glove, and the object mask is essential for preserving fine-grained contact details.

\begin{figure}
  \centering
   \includegraphics[width=0.8\linewidth]{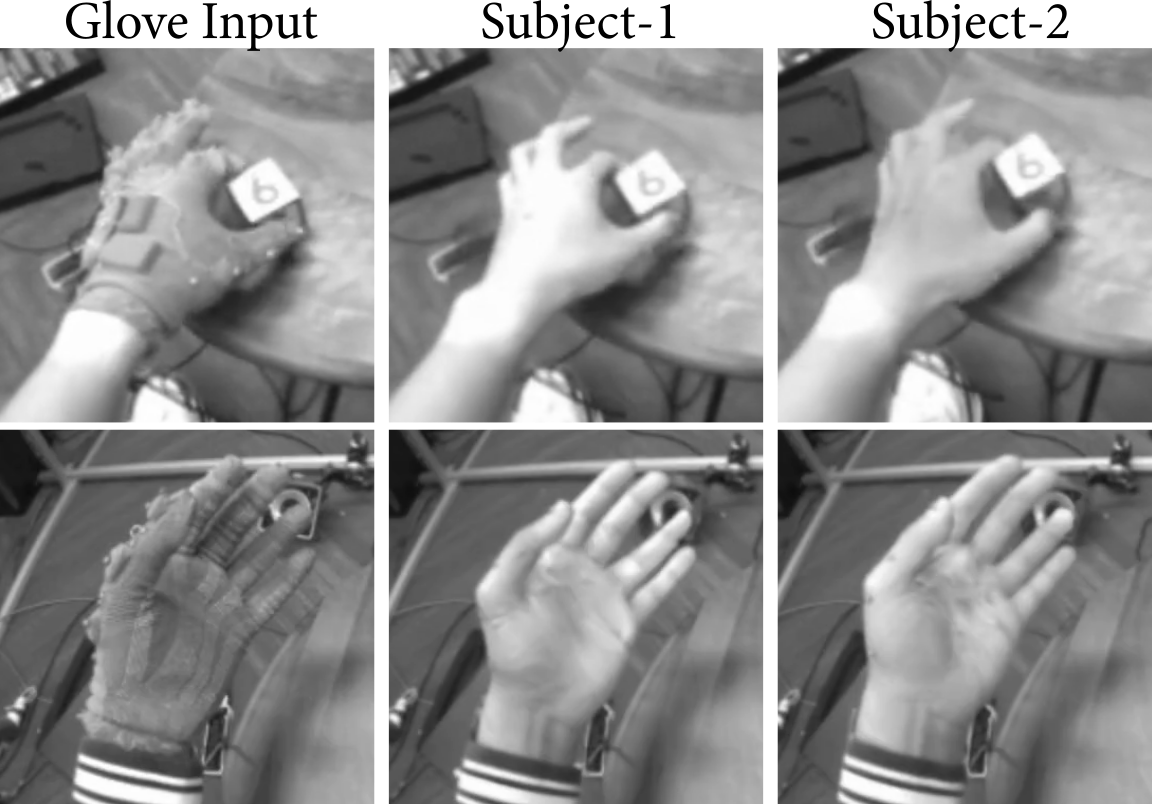}
   \caption{\textbf{Glove-to-Hand (different subjects).} We customize the output hand identity by training a 3D Gaussian hand per-subject, while sharing the diffusion hand restorer. The figure shows glove inputs (left) and the translated hand of different subjects (right).}
   \label{fig:cross-person}
\end{figure}



\subsection{Bare-Hand Applications}
\label{sec:exp-app}

\subsubsection{Vision-based Contact Estimation}
\label{sec:exp-downstream-1}

\noindent\textbf{Setup.} Estimating hand-object contact from vision is challenging, and there is a lack of large-scale, accurately-labeled data. Existing methods like PressureVision~\cite{grady2022pressurevision, grady2024pressurevision++} capture contact on planar surfaces~\cite{senselpad}, but this setup is insufficient for dexterous hand-object interactions. Our Glove2Hand framework addresses this by generating realistic hand videos with synchronized tactile signals, which serve as dense ground-truth contact labels. We leverage this data to train a video-based contact estimator. The model takes a 32-frame clip as input and predicts binary contact states for each fingertip at each frame, yielding an output of size $\mathbb{R}^{32\times5}$. The architecture uses a frozen DINOv3 backbone~\cite{simeoni2025dinov3} fine-tuned with LoRA~\cite{hu2022lora}. We apply four temporal attention layers over the [CLS] tokens, followed by a linear classifier to predict contact logits. The model is trained on a combination of real hand videos and our generated glove-to-hand videos. Real hand videos are manually annotated, while ground-truth for our generated videos is obtained by thresholding the tactile signals.

\noindent\textbf{Results.} We evaluate all models on a held-out test set of manually annotated real hand videos in Tab.~\ref{tab:downstream-1} with using Contact IoU~\cite{grady2024pressurevision++}, precision, and recall. A model trained on the translated videos significantly outperforms the one trained on the original glove videos, while combining our data with real videos yields the best performance. This validates our framework as a data curation engine and provides strong evidence for the generation quality. 

\begin{table}\centering
\resizebox{\linewidth}{!}{
\begin{tabular}{ccccc}\toprule
\multicolumn{2}{c}{Method} &Contact IoU (\%) &Precision (\%) &Recall (\%) \\\midrule
\multicolumn{2}{c}{PressureVision++~\cite{grady2024pressurevision++}} &0.8 &62.5 &0.9 \\
\multirow{4}{*}{Ours} &Glove only &71.5 &82.8 &83.9 \\
&G2H only &75.6 &90.6 &82 \\
&Hand only &85.3 &90 &94.2 \\
&Hand+G2H &\textbf{88.2} &\textbf{92.6} &\textbf{94.9} \\
\bottomrule
\end{tabular}}
\caption{\textbf{Results on Contact Estimation.} We train video contact estimator on glove-to-hand (G2H) videos, and evaluate on real, manually-annotated videos using Contact IoU~\cite{grady2022pressurevision}, Precision, and Recall. Training on G2H videos significantly outperforms the glove ones. Combining real and G2H data yields the best result. G2H contact labels are automatically derived from tactile signals.}\label{tab:downstream-1}
\end{table}

\begin{figure}
  \centering
   \includegraphics[width=0.9\linewidth]{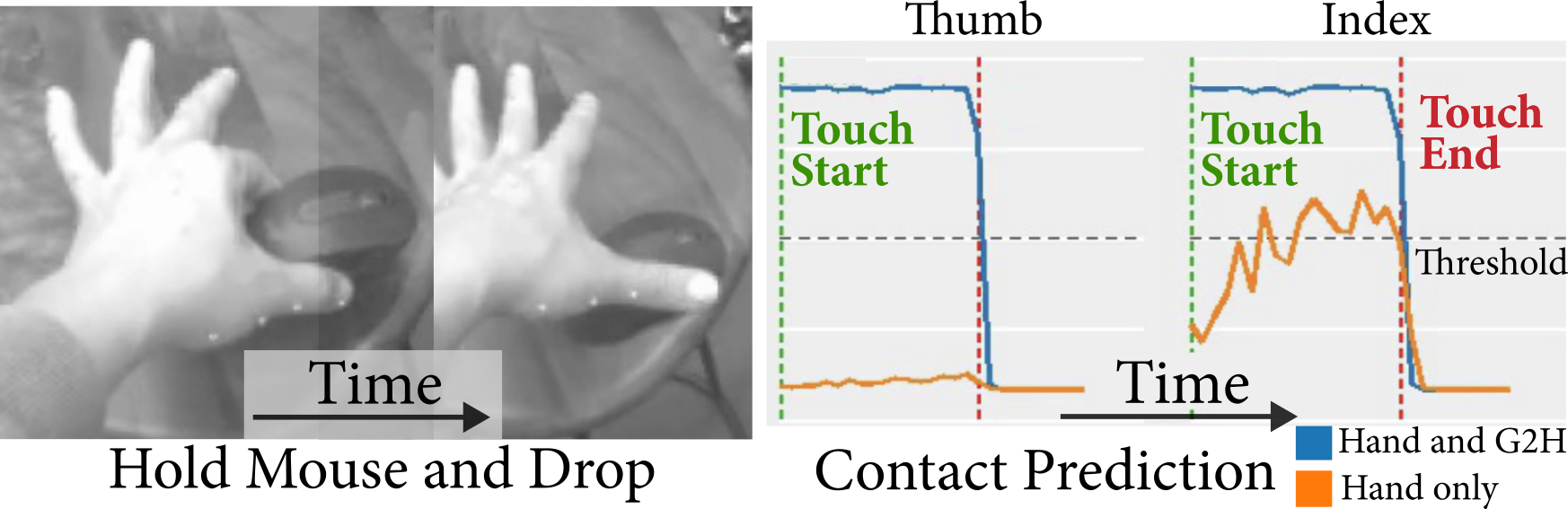}
   \caption{\textbf{Contact Estimation Example.} The model trained with only manually-annotated real hand data (\textcolor{orange}{orange}) produces noisy and unstable predictions. In contrast, adding our glove-to-hand (G2H) data (\textcolor{blue}{blue}) results in stable and accurate contact estimation.}
   \label{fig:downstream-1}
\end{figure}

\subsubsection{Hand Tracking under Heavy Occlusion}
\label{sec:exp-downstream-2}

\begin{figure}
  \centering
   \includegraphics[width=\linewidth]{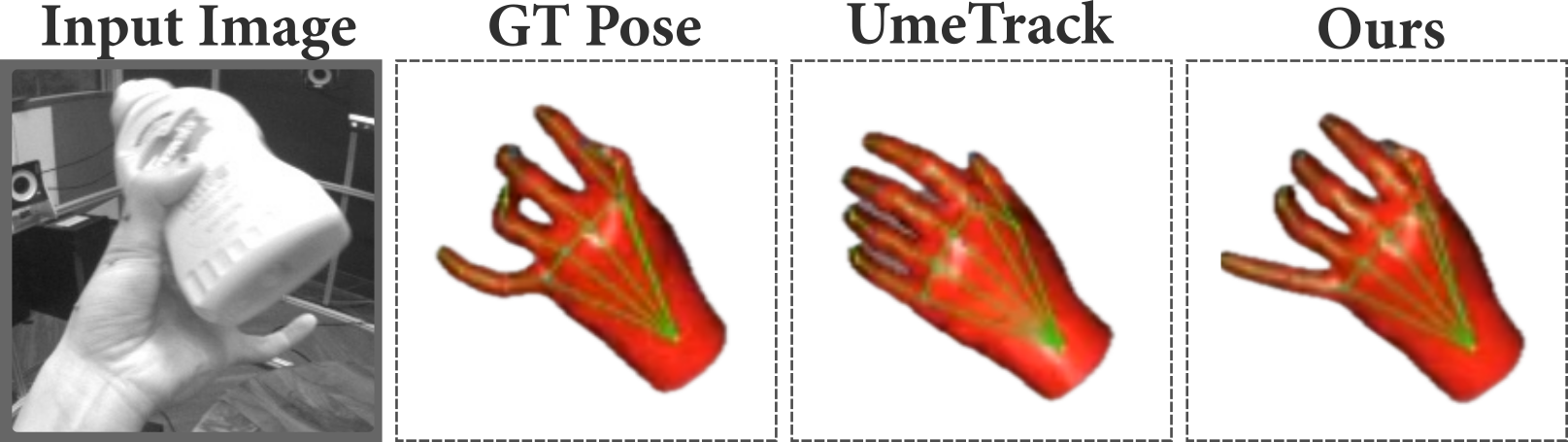}
   \caption{\textbf{Hand Tracking Visualization.}  UmeTrack struggles with occlusions, our model, finetuned on our synthesized videos with IMU tracking annotation, robustly tracks the occluded hand. }
   \label{fig:downstream-2}
\end{figure}
Estimating hand pose during object interaction is critical, yet existing methods are frequently limited by self- and object-occlusion. A key barrier is the difficulty of acquiring ground-truth pose annotations in such occluded scenarios~\cite{chao2021dexycb, hampali2020honnotate}.
We address this by leveraging on-glove IMU sensors. They provide accurate, vision-free joint measurements, enabling occluded pose capture without multi-camera studios which is suitable for in-the-wild recording.
To evaluate, we finetune UmeTrack~\cite{han2022umetrack}, a strong egocentric hand tracker, on our HandSense dataset using IMU-derived poses. We report the mean keypoint position error (MKPE) in mm, on a held-out test set of real videos with poses captured in a multi-camera MoCap system as ground truth. As shown in Tab.~\ref{tab:downstream-2}, finetuning with our glove-to-hand data significantly improves tracking accuracy, particularly in occluded cases (19.2 mm  $\rightarrow$  16.6 mm). Conversely, naive training on raw glove data degrades performance (19.5 mm $\rightarrow$ 26.5 mm), confirming a large domain gap. These results show that our framework can leverage IMU sensing to achieve robust hand pose tracking under severe occlusion.

\begin{table}\centering
\resizebox{\linewidth}{!}{ 
\begin{tabular}{lcccc}\toprule
\multirow{2}{*}{\textbf{Method}} &\multicolumn{2}{c}{\textbf{MKPE $\downarrow$}} &\multicolumn{2}{c}{\textbf{MKPE.T $\downarrow$}} \\\cmidrule{2-5}
&\textbf{Occlusion} &\textbf{Overall} &\textbf{Occlusion} &\textbf{Overall} \\\midrule
UmeTrack~\cite{han2022umetrack} &19.2 &19.5 &10.8 &9.8 \\
UmeTrack + Glove &27.2 &26.5 &11.5 &11.2 \\
UmeTrack + G2H (Ours) &\textbf{16.6} &\textbf{17.8} &\textbf{9.9} &\textbf{9.4} \\
\bottomrule
\end{tabular}}

\caption{\textbf{Results on Hand Tracking.} We directly evaluate UmeTrack on HandSense and finetune it with IMU tracking results as ground-truth. Finetuning with Glove2Hand videos improves tracking accuracy over UmeTrack, especially for occluded cases as IMU sensors are robust to camera occlusion. In contrast, training on glove data degrades performance due to the domain gap.}\label{tab:downstream-2}
\end{table}
\section{Conclusion}
We present the Glove2Hand framework for synthesizing photorealistic bare-hand videos from sensor-glove videos, capable of handling complex object interactions while preserving synchronization with multi-modal sensor signals. 
Leveraging Glove2Hand, we construct HandSense, the first multi-modal HOI dataset and demonstrate its effectiveness in significantly enhancing bare-hand applications.
We believe our work enables more physically grounded and realistic analysis of hand-object interactions, benefiting broader applications in the computer vision community.

{
    \small

}

\clearpage
\appendix

\begin{figure*}[t]
  \centering
   \includegraphics[width=\linewidth]{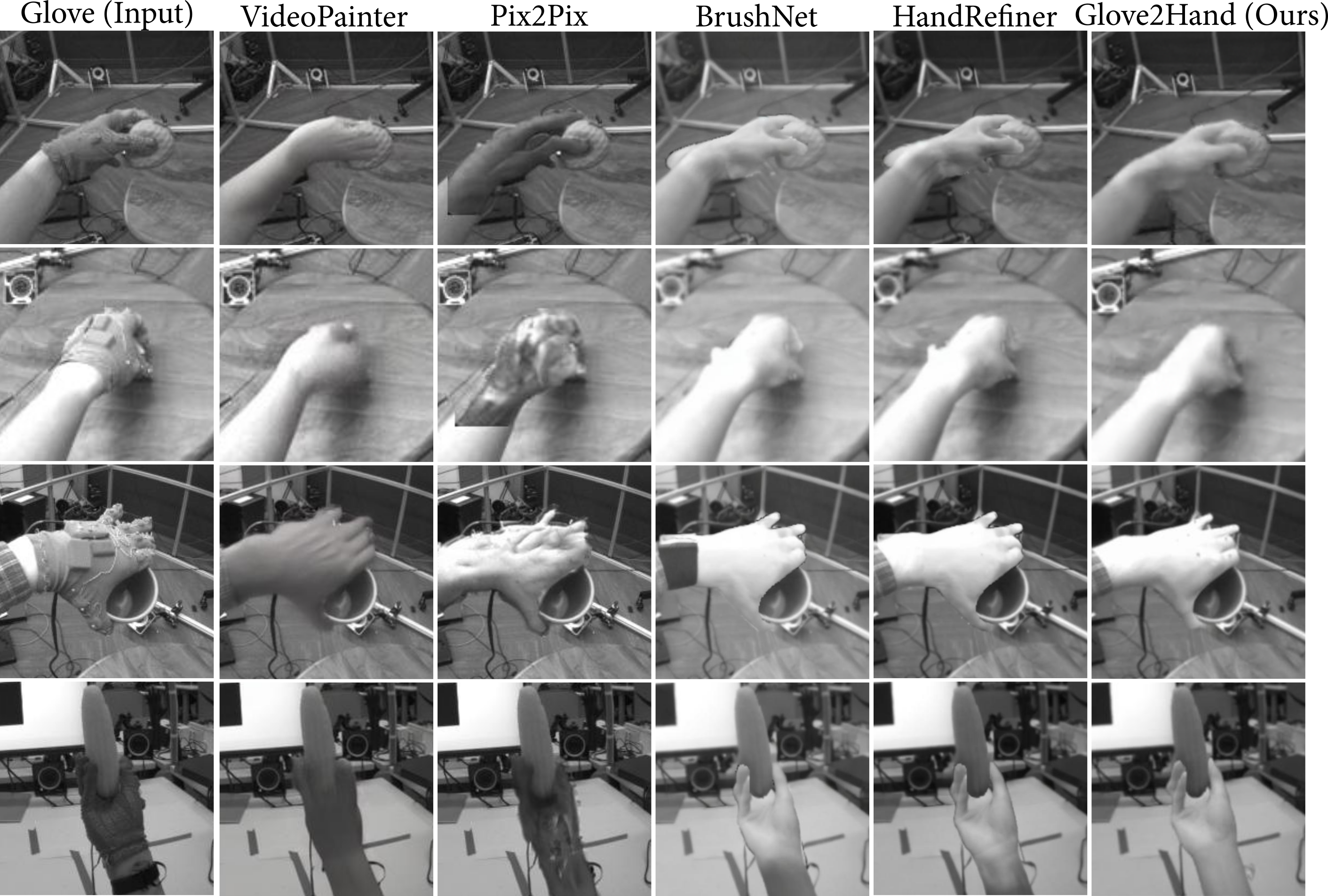}
   \caption{\textbf{Additional Qualitative Comparison for Glove-to-Hand.}}
   \label{fig:qualitative-add}
\end{figure*}

\begin{center}
    \interlinepenalty=10000
    {\Large \bf Appendix \par} 
\end{center}
In this appendix, we first introduce our sensing glove configuration. Next, we outline the data collection protocol for the \textit{HandSense} dataset. We then detail the our user study process. Finally, we provide implementation details regarding the training and inference of the Glove2Hand framework. Fig.~\ref{fig:qualitative-add} provides additional qualitative comparison with existing work. Please refer to the supplementary video \texttt{glove2hand-visualization.mp4} for visualizations of synthesized hand-object interactions.

\section{Sensing Glove Configuration}
\label{sec:glove_config}

Although Glove2Hand is general to handle various sensor glove design, the glove chosen in this paper is our research platform designed for advanced hand sensing, tracking, and haptic feedback in AR/VR/MR applications. It features 12 IMUs for detailed hand pose estimation and 5 capacitive tactile sensors on the fingertips for touch and pressure detection, supporting microgestures and interactions with physical objects. The glove streams sensor data at high framerates suitable for real-time gesture recognition and hand tracking. More detailed information about the glove configuration will be shared upon acceptance of the paper to comply with the anonymity requirements of the double-blind review.

\section{HandSense Data Collection}
\label{sec:data_collection}

We collect data spanning six hand-object interaction tasks, summarized in Table~\ref{tab:tasks}.

\begin{table*}[h]
\centering
\caption{List of interaction tasks and exemplar instructions used during data collection.}
\label{tab:tasks}
\resizebox{\linewidth}{!}{%
\begin{tabular}{ll}
\toprule
\textbf{Task} & \textbf{Exemplar Instruction} \\
\midrule
1. Bottle Opening & ``Open the mustard bottle using only your index finger and thumb.'' \\
2. Object Rotation & ``Hold the object with index, thumb, and middle fingers; rotate it in front of you.'' \\
3. Piano Keystroke & ``Press different piano keys sequentially using only your ring finger.'' \\
4. Pick-and-Place & ``Pick up the object using a whole-hand grasp and place it on the table.'' \\
5. Surface Interaction & ``Press and slide your index finger firmly against the table surface.'' \\
6. In-Hand Rotation & ``Rotate the object within your hand, maximizing finger contact and occlusion.'' \\
\bottomrule
\end{tabular}
}
\end{table*}

The object set includes ten items: a mouse, phone, soda can, marker pen, piano key, squishy toy, cube, mug, table surface, and mustard bottle. For each subject, we record paired sessions—one wearing the tactile glove and one with a bare hand—using identical object configurations and task instructions. Data acquisition consists of 2--5 minute segments per task. Within each segment, subjects perform repeated trials while varying finger usage and grasp types. Prior to recording, retro-reflective markers are affixed to the dorsal surface of the subject's hand (or glove) to enable ground truth pose tracking via an optical motion capture system. We compare HandSense with other datasets with contact label in Tab.~\ref{tab:statistics}.

\noindent \textbf{Motion Retarget from Gloves to Bare Hands.} Although glove and human hands share the same kinematic structure, the glove is bulkier than the bare hand. Directly using the glove pose as a hand pose results in unrealistic glove2hand synthesis. For example, an OK sign collected from the glove, when synthesized to a bare hand, would show the index finger and thumb not touching due to the glove's surface thickness. We treat this as a per-joint displacement. To mitigate this, at the beginning of glove data collection, we ask each candidate to perform a set of standard hand poses, e.g., pinch, OK sign, while wearing the glove. We then optimize the joint displacements and apply them to each recorded glove hand pose. This process is analogous to motion retargeting in robotics where human motion is transferred to robot, but simpler because the gap between glove and bare hand is smaller.

\begin{table*}[!htp]\centering
\begin{tabular}{lcccc}\toprule
\textbf{Dataset} &\textbf{Images} &\textbf{Subjects/Objects} &\textbf{Pose} &\textbf{Contact} \\\midrule
H2O~\cite{kwon2021h2o} &572K &4 / 8 &Optim. &Estim. \\
ARCTIC~\cite{fan2023arctic} &2.1M &10 / 11 &MoCap &Estim. \\
HO-Cap~\cite{wang2024ho} &699K &9 / 64 &Optim. &- \\
HOI4D~\cite{liu2022hoi4d} &2.4M &4 / 800 &Manual &- \\
HOT3D~\cite{banerjee2025hot3d} &3.7M &19 / 33 &MoCap & \\
HandSense (Ours) &200K &5 / 10 &MoCap &Measured. \\
\bottomrule
\end{tabular}
\caption{\textbf{Datasets with Contact Labels.} HandSense is the first HOI video dataset to provide direct, sensor-measured contact.}\label{tab:statistics}
\end{table*}

\begin{figure*}
    \centering
    \includegraphics[width=\linewidth]{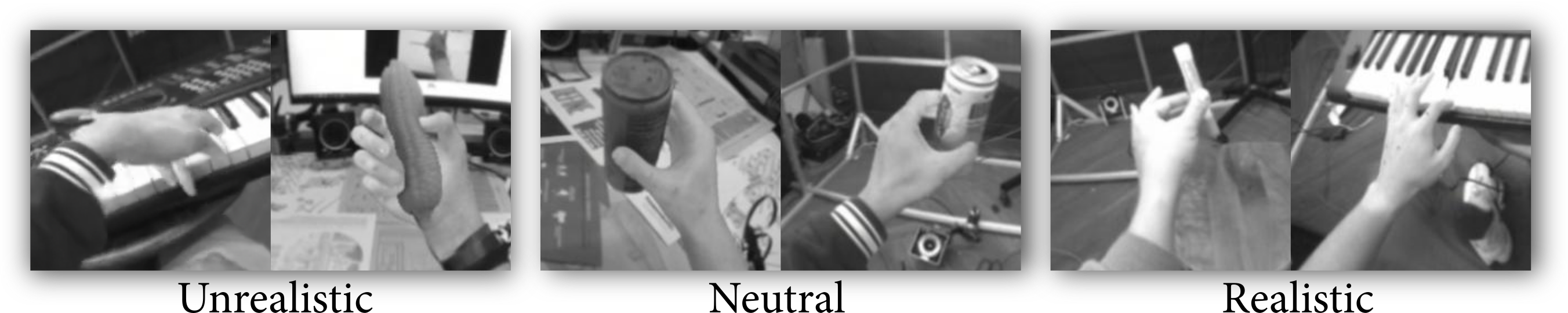}
    \caption{\textbf{Qualitative Examples of User Ratings.} Samples rated as ``Realistic'' or ``Very Realistic'' are perceptually indistinguishable from real hands. ``Neutral'' samples generally exhibit valid geometry and interaction plausibility but may lack skin details. ``Unrealistic'' samples typically display unnatural hand-object boundary or non-negligible visual artifacts.}
    \label{fig:userstudy-case}
\end{figure*}

\section{Human Evaluation Details}
\label{sec:user_study}


\vspace{0.25em}\noindent We recruited five participants to evaluate the perceptual quality of our synthesized imagery. Each participant completed a 45-minute session, evaluating approximately 40 images and 40 videos. The evaluation protocol uses a 5-point Likert scale (with 1.0 intervals) to assess hand realism, hand-object interaction (HOI) realism, motion stability, identity consistency, and visual artifacts. The complete questionnaire is detailed in Fig.~\ref{fig:questions}. Representative samples for different rating categories are shown in Fig.~\ref{fig:userstudy-case}. Our user study was conducted under a protocol approved by the Institutional Review Board (IRB). All participants provided informed consent.

\noindent\textbf{Evaluation Protocols.} We evaluate Glove2Hand across three distinct scenarios to assess generalization capabilities:
\begin{enumerate}
    \item \textbf{In-Domain:} Glove and bare hand identities match (Subject A's glove $\to$ Subject A's bare hand). The subject, background, and objects were seen during training. This setup validates the method's capacity for controlled data generation.
    \item \textbf{Cross-Subject:} Glove and bare hand identities differ (Subject A's glove $\to$ Subject B's bare hand). The environment is seen, but the target hand morphology is synthesized from a different source subject.
    \item \textbf{In-the-Wild:} A fully unseen setting where subjects, objects, and backgrounds were not present in the training set. This represents the most challenging scenario for scalable data collection.
\end{enumerate}

\vspace{0.25em}During evaluation, samples from these three groups are randomly interleaved with real ground-truth data in a blind study design. This establishes a high-quality reference anchor (upper bound) for the ratings.

\vspace{0.25em}\noindent\textbf{Results.} Table~\ref{tab:user-study} reports the Mean Opinion Score (MOS) and the perceptual gap (difference) between synthesized and real data. Higher MOS indicates better quality, while a lower gap indicates higher fidelity to the ground truth. We observe that Glove2Hand achieves high fidelity in controlled settings (In-Domain and Cross-Subject), making it suitable for large-scale data curation. While performance degrades in the challenging In-the-Wild setting, the results remain respectable. We hypothesize that scaling the training dataset size and improving HOI segmentation masks will further bridge the domain gap.

\begin{table*}\centering
\begin{tabular}{lcccc}\toprule
\multicolumn{2}{c}{\multirow{2}{*}{\textbf{Metric}}} &\multicolumn{3}{c}{\textbf{Mean Opinion Score $\uparrow$ / Gap to Real $\downarrow$}} \\\cmidrule{3-5}
& &\textbf{In-Domain} &\textbf{Cross-Subject} &\textbf{In-the-Wild} \\\midrule
\multirow{2}{*}{Image} &Hand Realism &\textbf{4.04 / 0.02} &3.67 / 0.39 &2.68 / 1.37 \\
&HOI Realism &\textbf{4.06 / 0.01} &3.51 / 0.56 &2.61 / 1.46 \\ \midrule
\multirow{5}{*}{Video} &Hand Realism &\textbf{3.83 / 0.51} &3.65 / 0.69 &2.69 / 1.65 \\
&HOI Realism &\textbf{3.96 / 0.35} &3.78 / 0.53 &2.84 / 1.47 \\
&Motion Stability &\textbf{3.64 / 0.50} &3.37 / 0.77 &2.54 / 1.60 \\
&Identity Consistency &\textbf{2.70 / 0.26} &2.54 / 0.43 &2.20 / 0.79 \\
&Visual Artifacts &\textbf{3.22 / 0.51} &3.00 / 0.73 &2.19 / 1.53 \\
\bottomrule
\end{tabular}
\caption{\textbf{Human Evaluation Results.} We report the Mean Opinion Score (MOS) and the perceptual gap relative to real data (Gap). \textbf{In-Domain} synthesis achieves performance near ground truth (Gap $< 0.05$ for images). \textbf{Cross-Subject} synthesis maintains high efficacy with gaps consistently below 1.0. \textbf{In-the-Wild} performance reflects the expected challenge of unseen environments but remains within a reasonable qualitative range.}
\label{tab:user-study}
\end{table*}

\begin{figure*}
    \centering
    \includegraphics[width=0.8\linewidth]{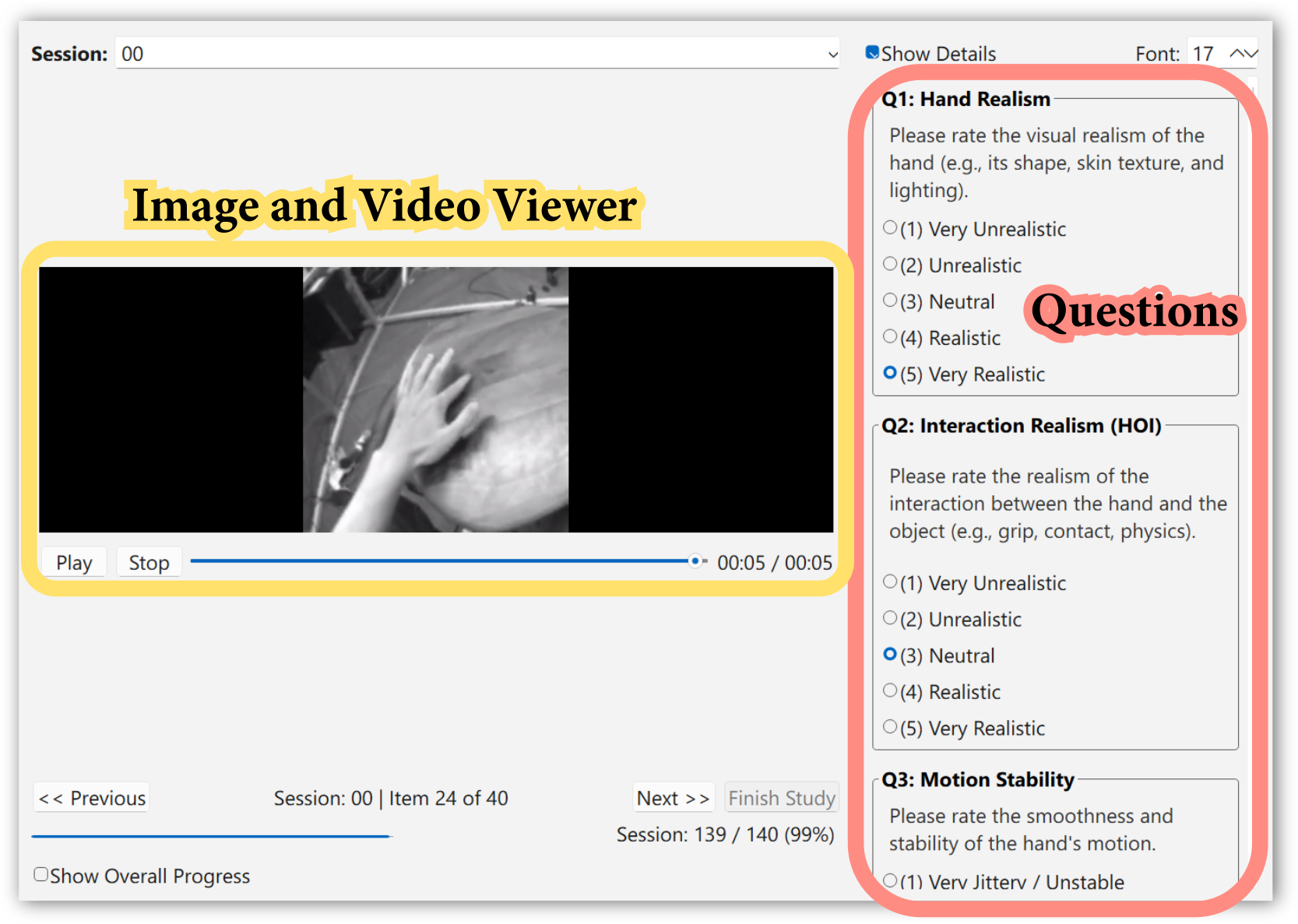}
    \caption{\textbf{Human Evaluation Interface.} Participants view a randomly sampled image or video and rate it against specific criteria before proceeding.}
    \label{fig:userstudy}
\end{figure*}

\begin{figure*}[t]
\begin{surveybox}
\footnotesize
\textbf{Per-Image Assessment}
\begin{enumerate}[label=\textbf{Q\arabic*:}, leftmargin=*, itemsep=0.5em]
    \item \textbf{Hand Realism:} Rate the visual realism of the hand (shape, skin texture, lighting).
    \begin{enumerate*}[label={(\arabic*)}, itemjoin={\hspace{1em}}]
        \item Very Unrealistic \item Unrealistic \item Neutral \item Realistic \item Very Realistic
    \end{enumerate*}
    
    \item \textbf{HOI Realism:} Rate the plausibility of the interaction (grip, contact, physics).
    \begin{enumerate*}[label={(\arabic*)}, itemjoin={\hspace{1em}}]
        \item Very Unrealistic \item Unrealistic \item Neutral \item Realistic \item Very Realistic
    \end{enumerate*}
\end{enumerate}

\vspace{0.5em}
\hrule
\vspace{0.5em}

\textbf{Per-Video Assessment}
\begin{enumerate}[label=\textbf{Q\arabic*:}, leftmargin=*, itemsep=0.5em, start=3]
    \item \textbf{Hand Realism:} Rate the visual realism of the hand.
    \begin{enumerate*}[label={(\arabic*)}, itemjoin={\hspace{1em}}]
        \item Very Unrealistic \item Unrealistic \item Neutral \item Realistic \item Very Realistic
    \end{enumerate*}
    
    \item \textbf{HOI Realism:} Rate the plausibility of the interaction.
    \begin{enumerate*}[label={(\arabic*)}, itemjoin={\hspace{1em}}]
        \item Very Unrealistic \item Unrealistic \item Neutral \item Realistic \item Very Realistic
    \end{enumerate*}
    
    \item \textbf{Motion Stability:} Rate the temporal smoothness of the hand motion.
    \begin{enumerate*}[label={(\arabic*)}, itemjoin={\hspace{1em}}]
        \item Very Unstable \item Unstable \item Neutral \item Stable \item Very Stable
    \end{enumerate*}
    
    \item \textbf{Identity Consistency:} How consistent is the hand's appearance (shape/size) over time?
    \begin{enumerate}[label={(\arabic*)}, nosep, leftmargin=*]
        \item Significant unnatural changes.
        \item Slight unnatural changes.
        \item Consistent appearance.
        \item (Unsure)
    \end{enumerate}
    
    \item \textbf{Visual Artifacts:} Are there distracting visual artifacts (blur, flickering, texture issues)?
    \begin{enumerate}[label={(\arabic*)}, nosep, leftmargin=*]
        \item Frequent/Severe artifacts.
        \item Noticeable artifacts.
        \item Minor artifacts.
        \item No artifacts observed.
    \end{enumerate}
\end{enumerate}
\end{surveybox}
\caption{\textbf{Human Evaluation Questionnaire.}}
\label{fig:questions}
\end{figure*}

\section{Glove2Hand Details}
\label{sec:imp_details}

\noindent\textbf{HOI Segmentation Masks.} To generate segmentation masks for hands and interacting objects, we implement a pipeline leveraging Grounding DINO~\cite{liu2024grounding} and SAM-2~\cite{ravi2024sam}. First, we detect potential objects using Grounding DINO. To identify the specific object being manipulated, we compute the Intersection-over-Union (IoU) between the detected object bounding boxes and the rasterized projection of the fitted hand mesh. The object with the highest IoU (surpassing a valid threshold) is selected as the box prompt for SAM-2. 
For hand segmentation, we detect the full arm using Grounding DINO and use the resulting box to prompt SAM-2. To ensure temporal consistency, we initialize SAM-2 with prompts on a single reference frame and propagate the masks. Finally, the specific hand or glove mask is obtained by cropping the full arm mask using the bounding box of the projected hand mesh.

\vspace{0.25em}
\noindent\textbf{Pose Optimization.} While we utilize an optical motion capture system, the ground truth hand pose $\mathbf{P}$ may exhibit inaccuracies due to marker occlusion or synchronization latency during rapid motion. To mitigate this, we introduce a learnable per-frame pose refinement term $\Delta \mathbf{P}$. This offset is optimized jointly with the Gaussian parameters during the reconstruction phase, following the camera pose optimization strategy in gsplat~\cite{ye2025gsplat}. We empirically find this refinement significantly reduces artifacts in the reconstructed Gaussian hand. Note that $\Delta \mathbf{P}$ is not used during the subsequent training of the diffusion restorer.

\vspace{0.25em}
\noindent\textbf{Gaussian Parameterization.} We anchor the 3D Gaussians to the mesh surface using barycentric coordinates. During optimization, we learn the unnormalized barycentric logits rather than the weights directly to ensure valid constraints. Additionally, we learn a scalar offset along the surface normal. This offset is parameterized via a sigmoid activation scaled by a hyper-parameter $z_{\text{max}}$, ensuring the Gaussians remain tightly grounded to the underlying geometry. Each subject-specific Gaussian model is trained on approximately 10 minutes of egocentric hand-only videos.

\vspace{0.25em}
\noindent\textbf{Auxiliary Training Data (HOT3D).} We incorporate the HOT3D dataset~\cite{banerjee2025hot3d} to augment the training of the Diffusion Hand Restorer. Unlike our primary subjects, we do not train 3D Gaussian models for HOT3D sequences. Instead, we employ a 2D self-supervised strategy: we crop out the hand region with a dilated (larger) hand mask, and mask out the wrist region before overlaying the original hand pixels onto the background. This creates videos of missing wrist details and hand-object boundary. The diffusion model is then trained to restore these corrupted regions (i.e., inpainting the hand-object boundary and wrist), allowing us to leverage large-scale data without expensive 3D reconstruction.

\vspace{0.25em}
\noindent\textbf{Training and Inference Efficiency.} All models are trained on NVIDIA A100 (80GB) GPUs. We crop hand regions from the raw headset footage at a resolution of $250\times250$. These crops are upsampled to $512\times512$ to satisfy the input constraints of the diffusion restorer's VAE. Furthermore, prior to training the 3D Gaussian hand model, we rectify the images and camera parameters to convert the raw fisheye distortion into a standard pinhole camera model.
For the 3D Gaussian hand, we train each subject-specific model for 120k iterations ($\sim$12 hours), though varying the schedule shows convergence at $\sim$6 hours. Rendering speed is approximately 50 FPS without custom CUDA kernel optimization. 
For the diffusion hand restorer, training proceeds in two stages. First, the image-based restorer is trained for 60k iterations. Second, we insert AnimateDiff~\cite{guoanimatediff} motion adapters and fine-tune on 22-frame video clips for an additional 60k iterations. The total training time is approximately 72 hours. For long-video generation, we apply the temporal sliding window strategy from DiffuEraser~\cite{li2025diffueraser} to ensure consistency. The inference speed is approximately 0.5 FPS.


\end{document}